\documentclass{article}

\usepackage[preprint]{neurips_data_2022}

\usepackage[utf8]{inputenc} % allow utf-8 input
\usepackage[T1]{fontenc}    % use 8-bit T1 fonts
\usepackage{hyperref}       % hyperlinks
\usepackage{url}            % simple URL typesetting
\usepackage{booktabs}       % professional-quality tables
\usepackage{nicefrac}       % compact symbols for 1/2, etc.
\usepackage{microtype}      % microtypography
\usepackage[dvipsnames]{xcolor}         % colors
\usepackage{graphicx}
\usepackage{natbib}
\usepackage{tabularx}

\usepackage{amsfonts}
\usepackage{amsthm}
\usepackage{amsmath}

\usepackage{times}
\usepackage{wrapfig}
\usepackage[font=small,labelfont=bf]{caption}
\usepackage{booktabs}
\usepackage{enumitem}
\usepackage{subcaption}

\newcommand\blfootnote[1]{%
  \begingroup
  \renewcommand\thefootnote{}\footnote{#1}%
  \addtocounter{footnote}{-1}%
  \endgroup
}

\hypersetup{
  colorlinks   = true, %Colours links instead of ugly boxes
  urlcolor     = blue, %Colour for external hyperlinks
  linkcolor    = blue, %Colour of internal links
  citecolor   = red %Colour of citations
}

\title{Towards Understanding How Machines Can Learn Causal Overhypotheses}

\author{%
  Eliza Kosoy\textsuperscript{*,1},  David M. Chan\textsuperscript{*,1}, Adrian Liu\textsuperscript{1}, Jasmine Collins\textsuperscript{1}, Bryanna Kaufmann\textsuperscript{1},
  \And
  Sandy Han Huang\textsuperscript{2}, Jessica B. Hamrick\textsuperscript{2}, 
  John Canny\textsuperscript{1}, Nan Rosemary Ke\textsuperscript{2}, Alison Gopnik\textsuperscript{1} \\
}

\begin{document}

\blfootnote{ \textsuperscript{*} Authors contributed equally, \textsuperscript{1} University of California, Berkeley \textsuperscript{2} Deepmind,
Corresponding authors: \texttt{eko@berkeley.edu, davidchan@berkeley.edu}
}

\maketitle

% DON'T COMMENT THESE OUT!!!
\newcommand{\gh}{\textsc{given hypothesis}}
\newcommand{\ghconj}{\textsc{given hypothesis (conjunctive)}}
\newcommand{\ghdis}{\textsc{given hypothesis (disjunctive)}}
\newcommand{\ngh}{\textsc{not given hypothesis}}
\newcommand{\nghconj}{\textsc{not given hypothesis (conjunctive)}}
\newcommand{\nghdis}{\textsc{not given hypothesis (disjunctive)}}
\newcommand{\conj}{\textsc{conjunctive}}
\newcommand{\dis}{\textsc{disjunctive}}

\newcommand{\perstep}{\textsc{per-step}}
\newcommand{\minstep}{\textsc{minimum-step}}
\newcommand{\uniform}{\textsc{uniform}}
\newcommand{\experimental}{\textsc{experimental}}
%%%

% Formatting
%\setlist[itemize]{itemsep=0mm}

\begin{abstract}

Recent work in machine learning and cognitive science has suggested that understanding causal information is essential to the development of intelligence. The extensive literature in cognitive science using the ``blicket detector'' environment shows that children are adept at many kinds of causal inference and learning. We propose to adapt that environment for machine learning agents. One of the key challenges for current machine learning algorithms is modeling and understanding causal overhypotheses: transferable abstract hypotheses about sets of causal relationships. In contrast, even young children spontaneously learn and use causal overhypotheses. In this work, we present a new benchmark---a flexible environment which allows for the evaluation of existing techniques under variable causal overhypotheses---and demonstrate that many existing state-of-the-art methods have trouble generalizing in this environment. The code and resources for this benchmark are available at \url{https://github.com/CannyLab/casual_overhypotheses}.
\vspace{-1.5em}

\end{abstract}

% \blfootnote{Additional details and media are available at \url{https://cannylab.github.io/clear2022}}

\section{Introduction}
Over the past few years, deep learning has made impressive progress in many areas, including reinforcement learning, natural language processing, and computer vision. However, most state-of-the-art algorithms still pale in comparison to humans, even children, for out-of-distribution generalization and fast adaptation to new tasks.
In contrast, causal modeling has long concerned itself not merely with accurate modeling of in-distribution data, but also the accurate recovery of underlying causal mechanisms (and their true graphical relations) capable of explaining \textit{out-of-distribution} data. Hence causal modeling holds promise for achieving systematic generalization~\citep{bengio2019meta,scholkopf2021towards,ke2021systematic}.

Causal overhypotheses are a key component of causal learning---they enable humans to learn causal models from a sparse amount of data~\citep{griffiths2009causal}, because they reduce the set of possible causal relationships to consider. Despite the recent surge of machine learning datasets and environment for causal inference and learning~\citep{ahmed2020causalworld,mcduff2021causalcity,wang2021alchemy,ke2021systematic}, the causal overhypotheses for these environments and datasets are unclear. In this work, we evaluate existing machine learning algorithms for causal overhypothesis learning. Instead of proposing a new environment for learning causal overhypotheses, we draw inspiration from recent cognitive science development, and use a virtual blicket detector environment, which has been used for understanding causal overhypotheses learning in children~\citep{kosoy2022learning}.

A ``blicket detector'' is a machine that lights up and plays music when some combinations of objects but not others are placed on it ~\citep{gopnik2000blicket,lucas2014children}. The central question is whether an agent can learn that a particular set of causal events will lead to the lighting-up effect, and use that knowledge to design novel interventions on the machine. The causal relationship is entirely determined by the pattern of conditional dependencies and interventions, rather than requiring intuitive physics knowledge or visual understanding. Although these tasks may seem simple, and are easily mastered by children, they are challenging for current learning algorithms, as we will see later.

Several features of this environment and the tasks it allows make it particularly useful as a benchmark for machine learning algorithms. First, causal representations are more powerful and structured than mere statistical generalizations, though both can be systematically inferred from statistical information. Many researchers %(e.g. Pearl, Spirtes et al., Bengio)
have argued that such causal representations are crucial for both human and general AI intelligence. The goals of causal inference, namely learning which actions will alter the environment in particular ways, are similar to those of standard reinforcement learning, but adding causal representations and inferences makes such learning far more effective. 

Second, and unlike some existing causal environments \citep{ke2021systematic,wang2021alchemy}, the blicket environment enables the inference of both specific causal structure and more general features of causal structure, such as whether causal systems are conjunctive or disjunctive, stochastic or deterministic. Learning these ``overhypotheses'' about causal structure~\citep{griffiths2009causal} is especially important because such inferences can constrain the search for causal structure in the future, a search that can rapidly become unwieldy. Third, this environment allows for active learning and exploration in a way that is both sufficiently constrained to be practical and that also allows for informative interventions. 

Most significantly, and again unlike existing environments, research has already shown that even preschool children can easily manipulate and explore this environment, generate appropriate data, and rapidly learn both particular causal structure and overhypotheses about causal structure~\citep{kosoy2022learning}. We can then directly compare both the overall performance and the behavior of various agents in these tasks to the performance and behavior of children. Young children are a particularly informative baseline group. They do not have the extensive education and experience of typical adults, which might make comparisons to artificial agents challenging, but they are nevertheless effective causal learners and able to make broad yet accurate generalizations from small sample sizes, in contrast to many current machine learning systems~\citep{gopnik2012scientific,gopnik2017changes}.

In what follows, we look at a spectrum of algorithms ranging from reinforcement learning, to imitation learning, to the use of pre-trained language models, to better understand how these different methods perform on the proposed environment and tasks.
We find that such algorithms, in contrast to children, only converge on a solution after an extensive number of trials or if they are given all the possible settings and outcomes as training data. This suggests that these tasks are an interesting challenge for machine learning algorithms. In order for machines to perform as well as children do, algorithms must reason about the sequence of observations seen, extract causal overhypotheses from those observations and use them for exploration---which current methods fall short of doing.

\section{Related Work}
Our work relates to prior work on exploration in reinforcement learning (RL), multi-task RL, causal learning in RL, and large language models.

\paragraph{Exploration in Reinforcement Learning}
Recent exploration algorithms for deep reinforcement learning typically add an exploration bonus to the task reward; please refer to \citet{amin2021survey} for a comprehensive survey. This bonus could be based on novelty~\citep{bellemare2016unifying,ostrovski2017count,martin2017count,tang2017exploration,machado2018count}, dynamics prediction error~\citep{schmidhuber1991curious,pathak2017curiosity}, uncertainty~\citep{osband2016deep,burda2018exploration}, or disagreement~\citep{pathak2019self}. The vast majority of existing exploration methods do not involve causal reasoning, whereas our proposed environment requires algorithms to learn and use causal overhypotheses in order to explore effectively, to solve the task.

\paragraph{Multi-task and Causal RL Benchmarks}
There exist multi-task RL benchmarks featuring robotics \citep{yu2019meta,james2020rlbench}, physical reasoning \citep{bakhtin2019phyre,allen2019tools}, and video games \citep{cobbe2018quantifying,machado2018revisiting,nichol2018gotta,chevalier2018babyai}. Unfortunately, since it is not clear what the relevant causal overhypotheses for these environments are, it is difficult to evaluate how causal information influences agents' exploration.

RL benchmarks for causal discovery include Causal World \citep{ahmed2020causalworld}, Causal City \citep{mcduff2021causalcity}, Alchemy \citep{wang2021alchemy}, ACRE \citep{zhang2021acre}, and the work of \cite{ke2021systematic}. However, many of these environments either lack clear causal hypotheses or do not allow for controlling overhypotheses.
In addition, these environments primarily focus on causal induction or generalization, rather than exploration \citep[though see][]{sontakke2021causal}. In contrast, the blicket environment in this work is designed to measure agents' ability to explore using causal overhypotheses. Moreover, children have not been tested on any of these existing environments, whereas in the blicket environment, prior work has shown that children as young as age four are able to learn causal overhypotheses and use these to explore effectively~\citep{kosoy2022learning}. It can be informative to compare the exploration and performance of RL approaches to that of children.

\paragraph{Language models for reasoning tasks}
Large language models such as GPT \citep{radford2019language,brown2020language} and PALM \citep{chowdhery2022palm} are trained on massive amounts of data, and they have been shown to be able to express uncertainty and perform common sense reasoning up to an extent \citep{lin2022teaching}. 
In this work, we probe the the causal reasoning capabilities of GPT-3 and PALM using textual descriptions of the virtual blicket environment.

\section{The ``Blicket Detector'' Environment}
\label{sec:environment}

\begin{wrapfigure}{r}{0.33\textwidth}
\centering
    \vspace{-1em}
    \includegraphics[width=0.28\textwidth]{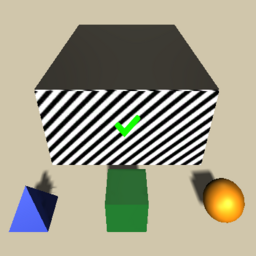}
    \caption{A screenshot of the embodied version of the blicket detector.} 
    \label{fig:blicket_detector}
    \vspace{-2\baselineskip}
\end{wrapfigure}

The blicket detector is commonly used in cognitive science to study causal inference and reasoning in children~\citep{gopnik2000blicket,lucas2014children}. Typically these studies are conducted in person. Recently \citet{kosoy2022learning} proposed an open-sourced virtual blicket detector environment (\autoref{fig:blicket_detector}), focused on studying how children learn and use causal overhypotheses in exploration. This virtual environment has several advantages: it enables not only tracking interactions with children, but also an \textit{exact} comparison between children and machine learning agents via the OpenAI Gym interface~\citep{1606.01540}. Please refer to \autoref{app:env} for examples of the virtual blicket detector in various states.

\paragraph{Causal Overhypotheses} The causal structure of this environment is hierarchical: the higher level specifies the causal relationship, and the lower level specifies which objects are blickets. Following \citet{kosoy2022learning}, we consider two causal relationships---disjunctive and conjunctive---which require that at least one or at least two blickets are placed on the detector in order for it to light up, respectively. More concretely, suppose there are three objects, $A$, $B$, and $C$, where $A$ and $B$ are blickets. In the conjunctive case, the detector would only light up if $AB$ or $ABC$ were placed. In the disjunctive case, any combination of objects except for just $C$ would cause the detector to light up.

\begin{figure}
\centering
    \includegraphics[width=\textwidth]{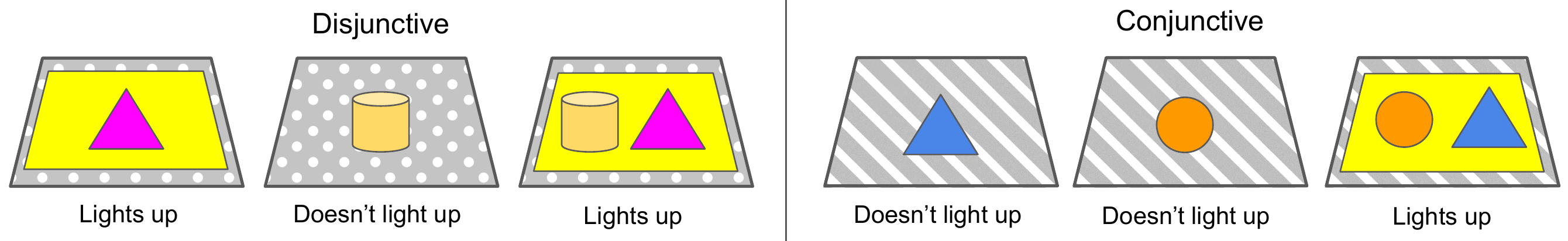}
    \caption{A simplified representation of the examples that children see in the \emph{given hypothesis} training phase.} 
    \label{fig:conj_disj}
\end{figure}

\paragraph{Prior Experiments with Children} Since we would like to compare the performance of machine learning agents to that of children in this environment, we aim to make our evaluation of machine learning algorithms comparable to how children are evaluated in \citet{kosoy2022learning}. Their evaluation used a $2\times 2$ experimental design, with a \emph{given hypotheses} or \emph{not given hypotheses} training phase, combined with a \emph{disjunctive} or \emph{conjunctive} causal relationship in the test phase. In the training phase, they showed the child two different blicket detectors, with three unique objects each, and for each detector demonstrated three examples of placing objects on the detector and checking whether it lights up. In the \emph{given hypotheses} case, one could infer from the examples that one detector was conjunctive and the other was disjunctive, as shown in \autoref{fig:conj_disj}. Whereas in the \emph{not given hypotheses} case, one could not infer from the examples whether the detectors were conjuctive or disjunctive. In the test phase, the children were given the opportunity to freely interact with a third blicket detector with three new objects. When they were done exploring, they were asked to respond yes or no to which objects were blickets. In their experiments with children from ages four to six, \citet{kosoy2022learning} found that children exhibit a diverse range of exploration strategies, which suggests that they are exploring based on a rich set of causal overhypotheses, formed from their prior knowledge of how objects and mechanisms behave.

\paragraph{Our Experiments with Models} To adapt the blicket environment for agent learning in this work, we made the following design choices regarding the observations, actions, reward, and termination conditions.
\begin{itemize}
    \item \textbf{Observations:} We could allow the algorithms to observe the same environment as the children, but this places RL algorithms at a significant disadvantage, since they would need to not only understand causal structures, but also learn visual inputs and object detection. Thus, we choose to evaluate the algorithms in a purely symbolic environment where the objects are represented by one-hot indices. 
    \item \textbf{Actions:} In the experiments with children, actions consist of placing a blicket onto the detector, removing a blicket from the detector, and pressing the ``check'' button to evaluate the detector's state. For RL algorithms, we simplify this process by allowing the agent to place multiple objects simultaneously, automatically ``checking'' the state of the detector with every action, and automatically resetting the detector after each check. This means that the agent gets feedback with every action, which significantly improved training stability.
    \item \textbf{Reward:} The reward function should capture whether the algorithm has learned the causal overhypothesis of an environment. To do this, we evaluate the models using a quiz-based framework. Models are allowed to make as many exploration steps as needed, and then trigger an action which switches to the evaluation mode. In the evaluation mode, models receive as input a blicket and must produce an action indicating if the object is a blicket or not. They receive a reward of $1$ for identifying a correct blicket, and a reward of $-1$ for incorrectly labeling an object (i.e., for both false positives and false negatives). We also explored several reward modifications. In one modification, models were asked to disambiguate between Disjunctive and Conjunctive environments, with +1 reward for identifying the correct modality.
\end{itemize}

The environment is implemented using the standard OpenAI Gym \citep{gym} interface, allowing it to be used across many different pre-existing machine learning architectures and algorithms. We hope that using this standard interface allows for expanded research in this field of causal overhypothesis learning.

% Section 4-------------------------------------------------------------------

%SECTION 5

\section{Benchmark Causal Learning in the Blicket Environment}
\label{sec:models_cmp}
Results from \cite{kosoy2020exploring} suggest that children can explore efficiently, especially given the causal overhypotheses. In this work, we evaluate how a spectrum of different machine learning models perform on the blicket detector tasks. Solving these tasks requires reasoning about the sequence of observations seen, extracting causal overhypotheses from those observations, and using these extracted overhypotheses for exploration.

We first evaluate several popular reinforcement learning algorithms---A2C~\citep{mnih2016asynchronous}, PPO2~\citep{schulman2017proximal}, and Q-learning~\citep{watkins1992q}---on this task. We also evaluate imitation learning algorithms, including behaviour cloning with decision transformers~\citep{chen2021decisiontransformer}. Finally, we apply pre-trained language models~\citep{brown2020language,chowdhery2022palm} to this task, since they have been shown to be capable of performing common-sense reasoning and expressing uncertainty to an extent \citep{lin2022teaching}.

\subsection{Deep Reinforcement Learning Algorithms}
\label{subsec:standard_rl}

We evaluate the performance of two pupular deep reinforcement learning algorithms, Advantage Actor Criric (A2C) \citep{mnih2016asynchronous} and Proximal Policy Optimization (PPO2) \citep{schulman2017proximal}, on the blicket environment. For each algorithm, we use several policy variants: a standard MLP policy (no memory), an LSTM-based policy, and an LSTM-based policy with layer normalization with two hidden dimension variants of 256 and 512. For all of these policies, we found that a network with a hidden size of 512 obtained the optimal performance See \autoref{app:hyper} for learning hyperparameters. We train all of these algorithms in the \textit{given hypothesis} scenario, where the agent is exposed during training to all possible overhypotheses, and asked to perform well given these scenarios.

\paragraph{Experimental Design} 
We terminate training either after 3 million environment steps or when the agent obtains maximum reward for 500 consecutive episodes, whichever comes first. Each episode has 25 timesteps, so each agent is exposed to at most 120,000 episodes. Experiments are run using a single Nvidia Titan-X Maxwell GPU, and take less than three hours to complete. To evaluate whether the agent can generalize to additional causal situations, we also train agents on five held-out scenarios: holding out all of the conjunctive overhypotheses, holding out all of the disjunctive overhypotheses, and holding out either one conjunctive overhypothesis or one disjunctive overhypothesis. 

\begin{figure}
    \centering
    \begin{subfigure}[t]{0.48\textwidth}
        \centering
        \includegraphics[width=\textwidth]{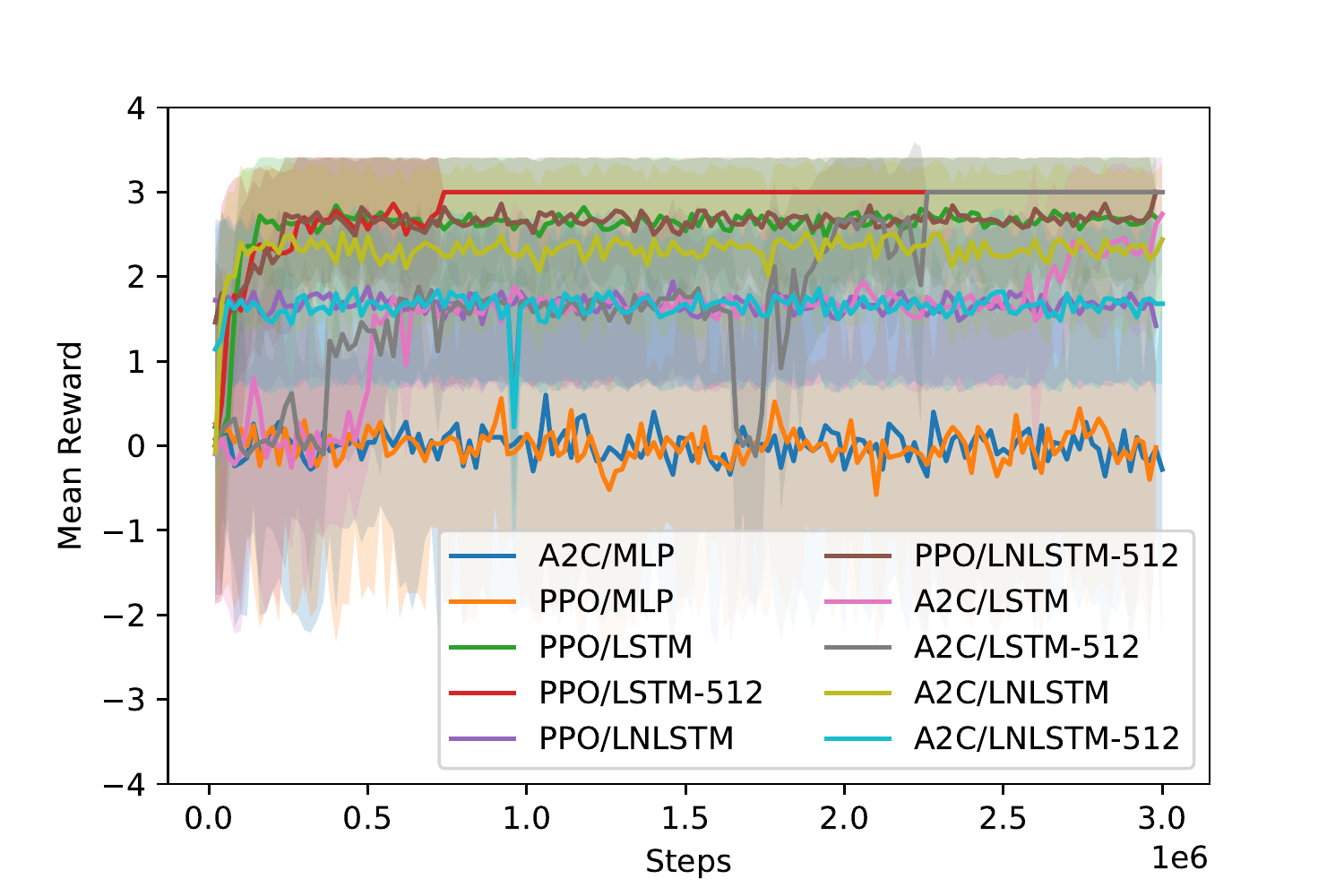}
        \subcaption{Performance of PPO2 and A2C algorithms with MLP, LSTM, and Layer-Norm LSTM policies on the blicket environment.}
        \label{fig:perf}
    \end{subfigure}
    \hfill
    \begin{subfigure}[t]{0.48\textwidth}
        \centering
        \includegraphics[width=\textwidth]{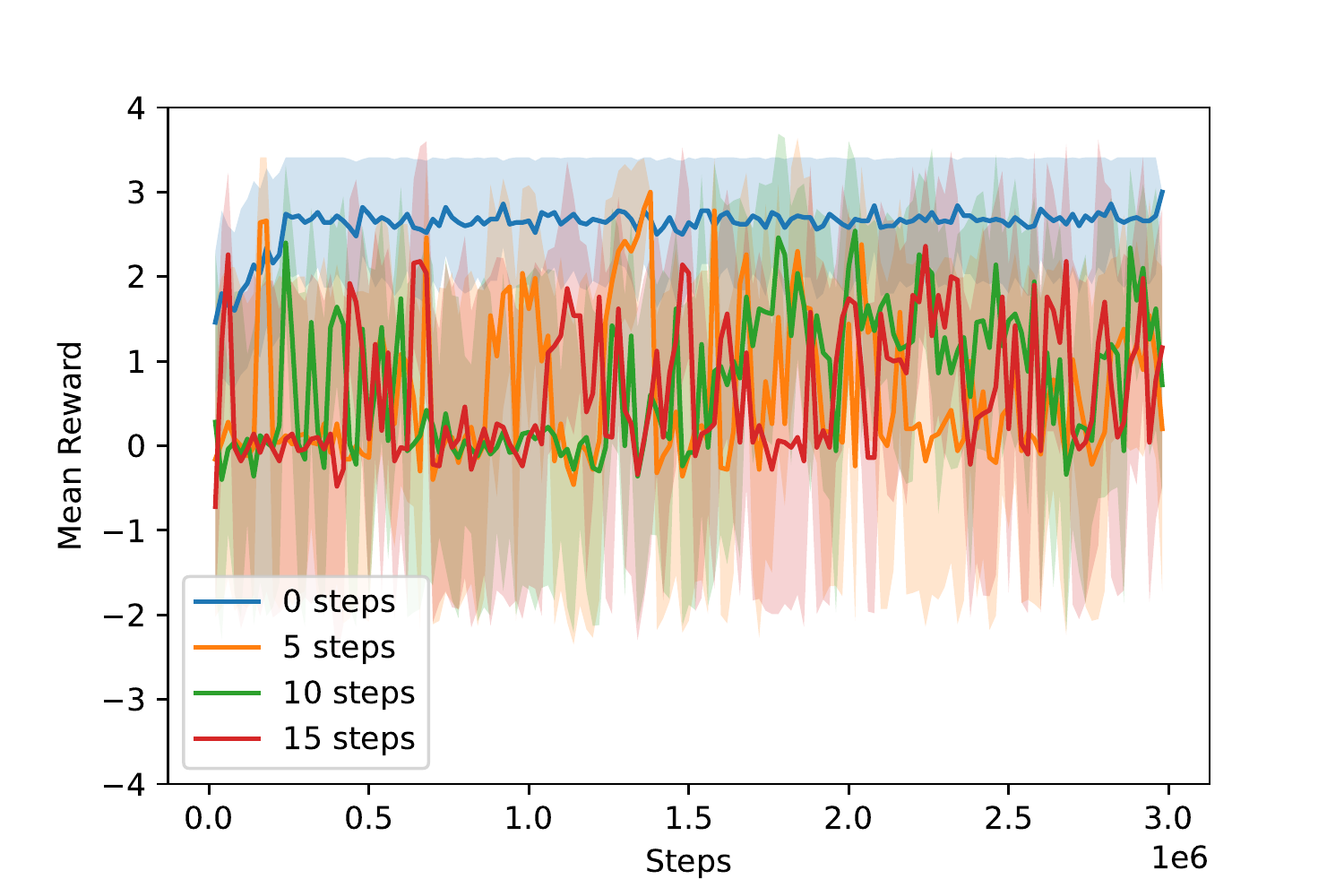}
        \subcaption{Performance when models are forced to explore for K steps before entering the quiz phase.}
        \label{fig:perf_forced}
    \end{subfigure}
    \\
    \begin{subfigure}[t]{0.48\textwidth}
        \centering
        \includegraphics[width=\textwidth]{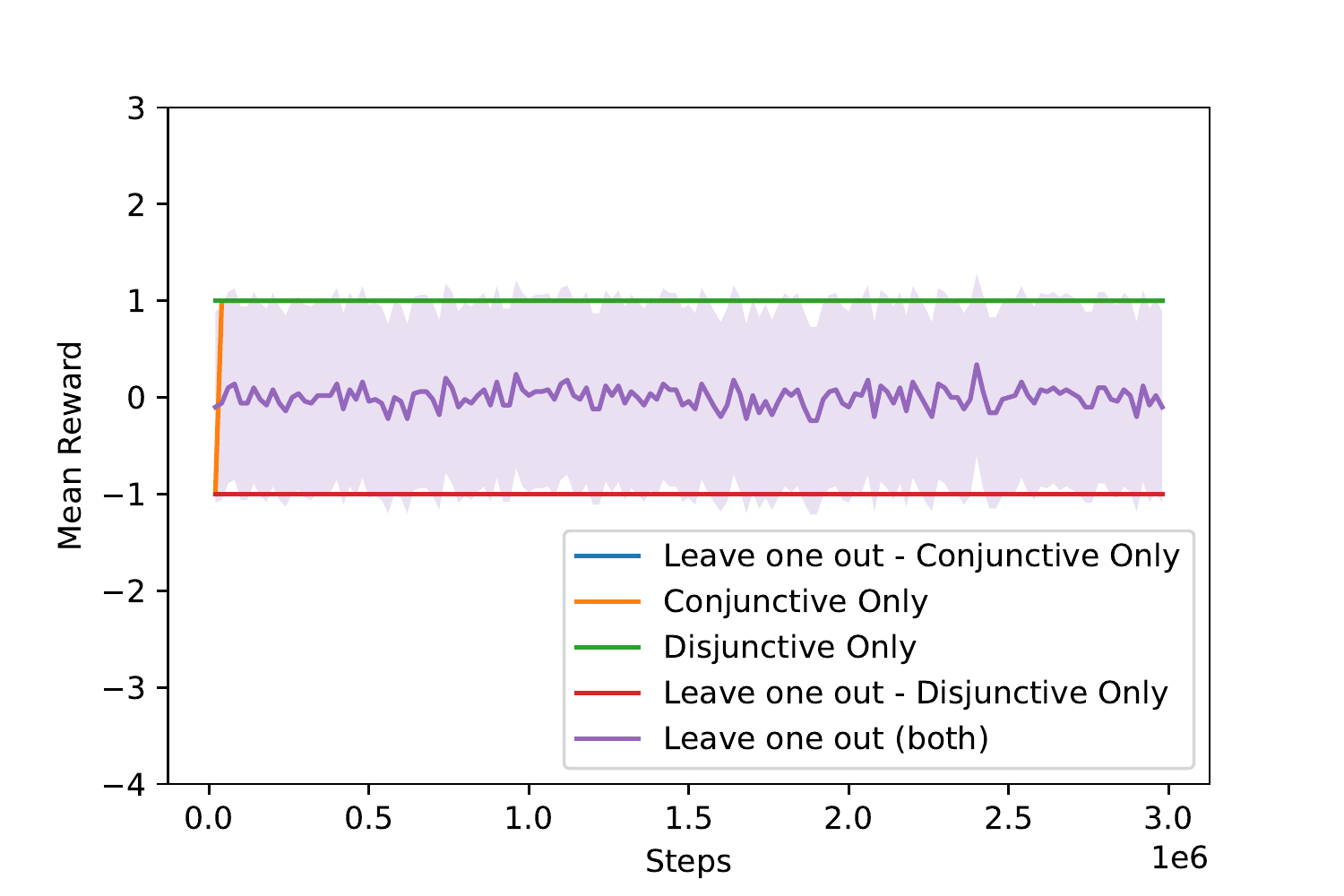}
        \subcaption{Performance on several over-hypothesis variants. Conjunctive Only: Train on only conjunctive hypotheses, test on both conjunctive and disjunctive. Disjunctive Only: Train on only disjunctive hypotheses, test on both conjunctive and disjunctive. Leave one out modifier: Train on 2/3 (or 4/6 in the case of both) of the hypothesis within the overhypothesis, and test on the remaining hypotheses.}
        \label{fig:perf_heldout}
    \end{subfigure}
    \hfill
    \begin{subfigure}[t]{0.48\textwidth}
        \centering
        \includegraphics[width=\textwidth]{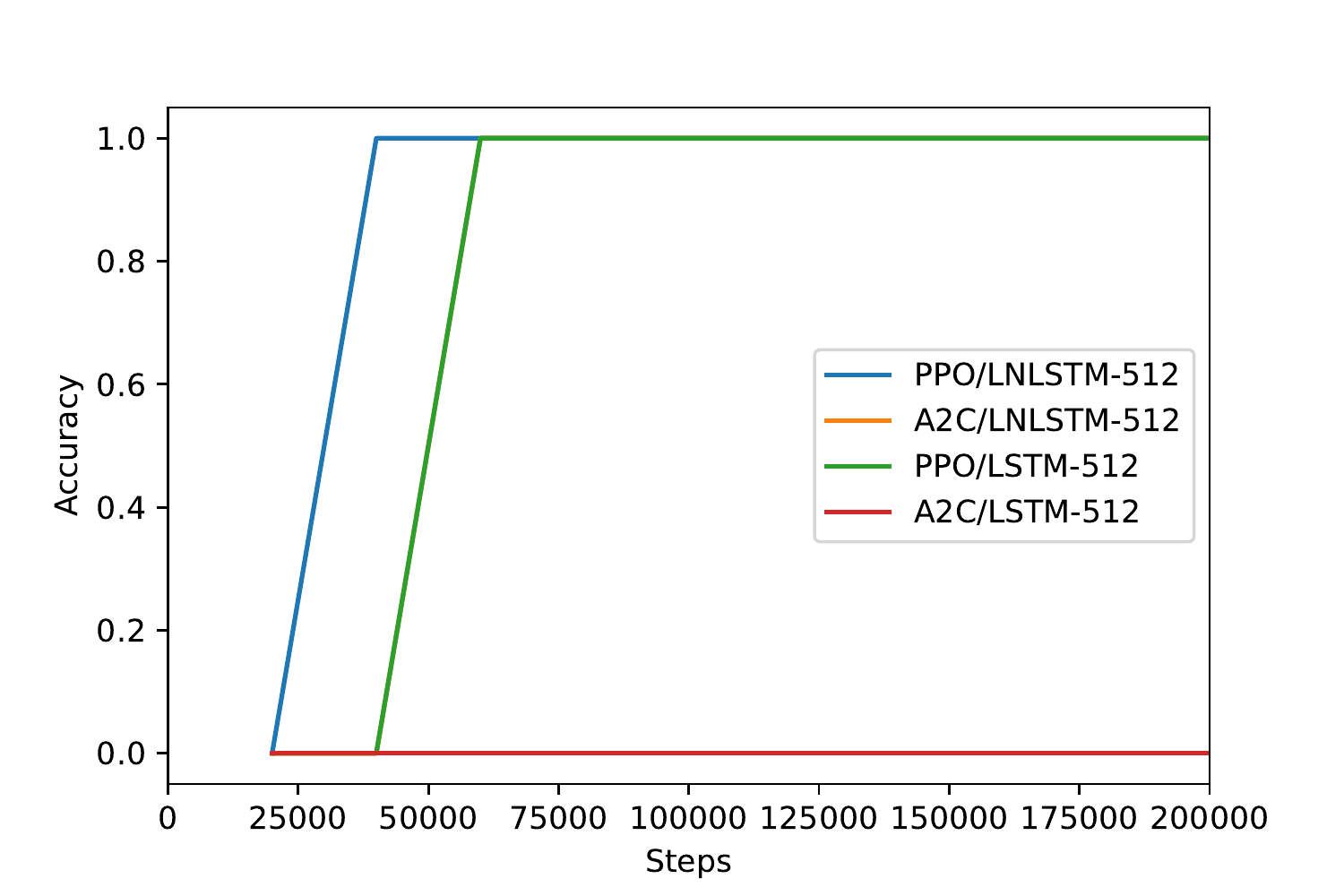}
        \subcaption{Accuracy of models over time when determining if the environment is conjunctive or disjunctive.}
        \label{fig:perf_acc}
    \end{subfigure}
    \caption{Experiments with standard deep reinforcement learning algorithms}
    \label{fig:standard}
\end{figure}

\paragraph{Results} \autoref{fig:perf} shows the performance of A2C and PPO2 when none of the hypotheses are held out. PPO2 outperforms A2C in almost all scenarios, achieving higher rewards faster. Further, the LSTM models clearly outperform the non memory-based models; this is expected, since causal learning requires memory. Unfortunately, none of the algorithms perform well on held-out causal examples, as shown in \autoref{fig:perf_heldout}. This suggests that they are primarily learning to memorize the causal patterns, and thus are incapable of generalization. This conclusion comes with a caveat: because there are only a handful of possible hypotheses, it may be possible that we do not have enough data to perform well on held-out samples. Agents are, however, able to easily distinguish between conjunctive and disjunctive environments---\autoref{fig:perf_acc} shows that after very few steps, in the held-out situation (with both overhypotheses), agents can distinguish conjunctive from disjunctive environments, even though they are unable to determine which objects exactly are blickets.

\subsection{Q-Learning}
\label{subsec:qlearning}
We also train tabular Q-Learning on the symbolic version of the blicket environment. We append the full history of previous observations to the current observation in order to give the agent memory. The Q-values are initialized to zero. We used $\epsilon$-greedy exploration with an exploration probability of $0.1$, and we found the best learning rate was $0.95$. Q-learning is able to learn the task very quickly due to a small search space---it took an average of $70$ episodes and $292$ steps for the agent to converge to maximum reward. However, tabular Q-learning is incapable of generalizing to new scenarios, so we do not test Q-Learning agents in the held-out scenarios.

\subsection{Behavior Cloning}

Recently \citet{chen2021decisiontransformer} introduced the Decision Transformer, a simple transformer-based approach to imitation learning, shown to outperform most existing behavior cloning methods. The decision transformer works by applying a causally-masked transformer to predict the reward-to-go of a flattened sequence of (state, action, reward, next state) tuples, with an $\ell^2$-norm loss. The model then chooses actions during test time that maximize the predicted reward-to-go. 

\paragraph{Experimental Design} In this work, we adapt the model to predict discrete actions in our space by adding a sigmoid activation to the action predictions and altering the action prediction loss accordingly. We evaluate the model using a target reward of 3, which corresponds to identifying all of the blickets correctly. The decision transformer was trained using a batch size of 128, a K-value of  30, embedding dimension of 128, 3 layers, one head, and dropout of 0.1. The weights are optimized using ADAM, with a learning rate of $1e^{-4}$ and weight decay of $1e^{-4}$. We also compare against standard behavior cloning, using an MLP with a hidden-dimension of 128, and trained using the same optimizer as the decision transformer. Both models are trained for 10 epochs using one Nvidia Titan-X Maxwell GPU, which takes less than one hour, and the best validation set checkpoint is used for each during test time. Models are evaluated on 100 rollouts in the environment, and the mean reward is reported.

When training the decision transformer model, we found that in some cases, random (and even the PPO2/A2C trained) models were unable to explore efficiently, as they entered the quiz environment too soon. Note that random exploration will enter the quiz environment after $t$ steps with with probability $p = 1 - \frac{1}{2^t}$. Thus, we found it helpful to force the policy to explore for several steps before allowing it to enter the quiz environment. While this process helps the decision transformer, it negatively affects the performance of the standard RL models, as shown in \autoref{fig:perf_forced}. 

\paragraph{Results} \autoref{tab:decision} shows the performance of the decision transformer and behavior cloning models when applied to datasets generated by several policies. As we can see, while expert policies allow for higher rewards using the decision transformer, allowing the model additional forced exploration time is the most important factor. This suggests that on their own, the A2C and PPO2 trained policies do not lead to sufficient exploration for learning a strong model of the reward, whereas forcing additional exploration (even if it is random) is much more useful. Notably, standard behavior cloning performs very poorly, as copying actions with the same local observations under different overhypotheses will likely lead to incorrect or uninformative actions.

\begin{table}
    \centering
    \small
    \begin{tabularx}{\linewidth}{lXcc}
        \hline
        \textbf{Model} & \textbf{Pre-Training Dataset} & \textbf{Reward} & \textbf{FCA}  \\
        \hline
        Decision Transformer & Random & $0.16 \pm 1.747$ & $0.67$ \\
        Decision Transformer & Random (5-step forced exploration) & $2.06 \pm 0.998$ & $0.71$ \\
        Decision Transformer & Random (10-step forced exploration) & $2.16 \pm 0.872$ & $0.73$ \\
        Decision Transformer & Random (15-step forced exploration) & $2.37 \pm 0.633$ & $0.77$ \\
        Decision Transformer & A2C-LSTM & $0.14 \pm 1.847$ & $0.63$ \\
        Decision Transformer & PPO2-LSTM & $0.26 \pm 0.990$ & $0.66$ \\
        Decision Transformer & A2C-LayerNorm-LSTM & $0.18 \pm 0.983$ & $0.62$ \\
        Decision Transformer & PPO2-LayerNorm-LSTM & $0.74 \pm 1.460$ & $0.64$ \\
        \hline
        Behavior Cloning & Random & $0.22 \pm 0.975$ & $0.61$ \\
        Behavior Cloning & Random (5-step forced exploration) & $-0.16 \pm 0.987$ & $0.58$ \\
        Behavior Cloning & Random (10-step forced exploration) & $0.14 \pm 0.990$ & $0.60$ \\
        Behavior Cloning & Random (15-step forced exploration) & $0.04 \pm 0.999$ & $0.61$ \\
        Behavior Cloning & A2C-LSTM & $0.10 \pm 0.995$ & $0.55$ \\
        Behavior Cloning & PPO2-LSTM & $0.06 \pm 0.998$  & $0.63$\\
        Behavior Cloning & A2C-LayerNorm-LSTM & $0.04 \pm 2.087$  & $0.59$\\
        Behavior Cloning & PPO2-LayerNorm-LSTM & $0.12 \pm 0.995$ & $0.68$ \\
        \hline
    \end{tabularx}
    \vspace{5pt}
    \caption{Performance of the imitation learning models with different pre-training datasets. Reward is on the blicket-quiz task: +1 for correctly identifying a blicket, and -1 for incorrectly identifying a blicket. FCA refers to Forced Choice Accuracy, the accuracy of the model in determining if the environment is conjunctive or disjunctive.}
    \label{tab:decision}
\end{table}

\subsection{Large Language Models}
\label{subsec:language_models}

Recently large language models have shown promising performance on a wide variety of tasks, including logical inference and common-sense reasoning. Such language models include GPT-3~\citep{brown2020language}, PaLM~\citep{chowdhery2022palm}, MT-NLG~\citep{smith2022mtnlg}, and Chinchilla~\citep{hoffmann2022chinchilla}, which all use transformer-based architectures and are trained autoregressively on a large corpus of text, to predict the next token when given a sequence of tokens. These language models have also shown promising performance on tasks that involve causal reasoning, such as identifying causal effects from observational data~\citep{veitch2020textcausal} and inferring causality for when drugs lead to adverse effects~\citep{wang2021inferbert} However, these models have not yet been applied to a task that involves reasoning across a set of overhypotheses, as is required in the blicket environment.

Motivated by this, here we investigate the performance of two such language models, GPT-3 and PaLM, on the blicket environment. One can apply a language model to a target task via purely text-based interaction with the model, by providing a prompt to the model and evaluating its output. This prompt could be \emph{freeform}, in which it has no set structure, or it could be \emph{few-shot}, where it contains a small number of examples of correct prompt-output pairs, followed by the test prompt. Instead of text-based interaction, another option is to fine-tune the model on the target task. This involves training the model on data from the target task, and can require a sizeable amount of data. To make the evaluation more similar to how children are introduced to the task, we focus on only text-based interaction with the model, including both freeform and few-shot prompts.

As in the experiments with children, there are four conditions: \emph{given} versus \emph{not given hypotheses}, and \emph{disjunctive} versus \emph{conjunctive}. In the \emph{disjunctive} case, one of the three objects is a blicket. In the \emph{conjunctive} case, two of the three objects are blickets. We evaluate how well the models can: 1) identify the blickets and 2) identify whether the causal structure is disjunctive or conjuctive. For GPT-3, we use the OpenAI API\footnote{Available at \url{https://beta.openai.com/overview}}, and for PaLM we use an internally available interface.

\paragraph{Freeform Prompt} In the freeform prompt, we provide the model with text that is as similar as possible to what children receive in the blicket experiment. The only difference is that we replace the visual components with text descriptions. We include the exact prompts in Appendix~\ref{sec:languageprompts}.  The prompt starts with a description of interacting with a striped blicket machine and then a dotted blicket machine, each of which has three unique objects. There are three interaction examples given per machine, for example ``If we put the blue pyramid on the machine, then it does not light up''. In the \emph{given hypotheses} condition, the striped blicket machine is conjunctive and the dotted blicket machine is disjunctive. In the \emph{not given hypotheses} condition, there is not enough information to determine whether the striped and dotted blicket machines are disjunctive or conjuctive. Thus the \emph{given hypotheses} condition defines the space of overhypotheses, whereas the \emph{not given hypotheses} condition does not. The prompt then introduces a new blicket machine, along with examples of interactions with the machine and whether it lights up or not. Finally we ask the model which objects are blickets and whether the new machine is more similar to the striped or dotted machine.

\paragraph{Few-shot Prompt} In the few-shot prompt, we structure the input into two prompt-output demonstrations, one per machine, containing the same information as is provided in the freeform prompt. Unlike the information given to children, in the outputs we explicitly state which objects are blickets. In addition, for the \emph{given hypotheses} condition, in the preamble we define the disjunctive and conjunctive hypotheses: ``A striped machine needs two blickets to make it light up, and a dotted machine needs one blicket to make it light up'', and in the outputs we state whether this is more similar to a striped or dotted machine. We need to structure the output like this because the output of the model will have the same structure.

The results are reported in Table~\ref{tab:language}.\footnote{For GPT-3, we use the text-davinci-002 model, with temperature 0.7, maximum length 256, and frequency and presence penalties 0. For PaLM, we use a temperature of 0 to obtain the greedy, or one-best decoding. When given the freeform input, the PaLM model output continues indefinitely, so we simply took the first two sentences of the output and ignored the rest.} We find that when given hypotheses, GPT-3 and PaLM are almost always able to select the correct causal structure, but they are not always able to select the correct blickets. In particular, GPT-3 frequently names too many objects as blickets, including those associated with other machines. In contrast, PaLM never identified objects as blickets that were not one of the three objects in the test task.

In the two-shot setting, PaLM performs best when the space of overhypotheses is covered perfectly by the two examples given, as one would expect. However, when this space is not covered, i.e. in the \emph{not given hypothesis} setting, PaLM struggles in the two-shot setting because the test example's causal structure does not match either of the two given examples. We tried adding chain-of-thought reasoning in the prompt for PaLM, but were not able to improve results for either the freeform or two-shot setting.

\begin{table}
    \centering
    \small
    \setlength\tabcolsep{5pt}
    \begin{tabularx}{\linewidth}{lXcc}
        \hline
        \textbf{Condition} & \textbf{Model \& Input} & \textbf{Blickets Chosen} & \textbf{Causal Structure}  \\
        \hline
        Given hypotheses, disjunctive & GPT-3, freeform & 1/1 correct, 6 wrong  & \textbf{correct} \\
        & PaLM, freeform & 1/1 correct, 1 wrong & wrong \\
        & PaLM, two-shot & \textbf{1/1 correct} & \textbf{correct} \\
        \hline
        Given hypotheses, conjunctive & GPT-3, freeform & 2/2 correct, 1 wrong & \textbf{correct} \\
        & PaLM, freeform & \textbf{2/2 correct} & \textbf{correct} \\
        & PaLM, two-shot & \textbf{2/2 correct} & \textbf{correct} \\
        \hline
        Not given hypotheses, disjunctive & GPT-3, freeform & 1/1 correct, 7 wrong & --- \\
        & PaLM, freeform & 0/1 correct & --- \\
        & PaLM, two-shot & 1/2 correct, 1 wrong & --- \\
        \hline
        Not given hypotheses, conjunctive & GPT-3, freeform & 2/2 correct, 7 wrong & --- \\
        & PaLM, freeform & \textbf{2/2 correct} & --- \\
        & PaLM, two-shot & 1/2 correct, 1 wrong & --- \\
        \hline
    \end{tabularx}
    \vspace{5pt}
    \caption{Results for GPT-3 and PaLM for the four conditions. Bold font indicates a fully correct answer. For the \emph{not given hypotheses} setting, there is not a correct answer possible for causal structure, because there is not enough information to determine whether the blicket detectors in the training phase are disjunctive or conjunctive.}
    \label{tab:language}
\end{table}

\section{Discussion \& Conclusion}
In this work, 
we looked into evaluating and understanding how machine learning models learn causal overhypotheses, by evaluating these models in the blicket environment. In contrast to existing benchmark tasks, in which there is a fixed causal structure, this environment focuses on the need for \emph{causal overhypotheses} in order to explore effectively to determine the underlying causal structure. We focused on three categories of state-of-the-art methods---deep RL, behavior cloning, and large language models---for tasks in this environment. In blicket detector experiments, children are able to learn causal overhypotheses from only a handful of observations and can apply these overhypotheses to explore effectively for a new situation~\citep{kosoy2022learning}. In contrast, our experiments indicate that state-of-the-art machine learning algorithms have difficulty learning and using causal overhypotheses for exploration and inference---we saw this in the weak performance of deep RL algorithms on held-out environments and in the tendency of decision transformer models to under-explore. In our experiments with language models, we provide the same observations that the children were given in \citet{kosoy2022learning}, along with a full set of examples for the new situation (thus removing the need for exploration). Despite this, language models struggle when the hypotheses are not given, and are not able to express uncertainty about the causal structure in that case.

Given that understanding and leveraging causal structure is essential to developing general intelligence, this work highlights an opportunity for improvement in this direction, and provides a set of concrete benchmark tasks to measure improvement. One direction of future work is to build machine learning models that can better learning causal overhypotheses. Modular architectures have shown to be helpful in understanding causal hypotheses of the environment \citep{goyal2019recurrent,goyal2021neural,ke2021systematic}; it would be promising to explore such models for causal overhypotheses understanding. Another direction of future work is to improve on exploration in RL agents by explicitly learning and incorporating causal overhypotheses, in order to narrow down the search over possibilities. Furthermore, an interesting direction is to train machine learning models on children's exploration trajectories from \cite{kosoy2020exploring}. This may lead to new insights for improving causal overhypotheses learning for machine learning models. A main limitation of this work is that we evaluated only two types of causal overhypotheses--conjunctive and disjunctive. These are particular types of overhypotheses on the causal graph structure. In the real world, there exist many other types of overhpotheses, as well as conditional probability distributions. It would be interesting to see if one can extend either the blicket or other RL environments (such as \cite{ahmed2020causalworld,ke2021systematic}) to include other types of causal overpotheses.

\section{Acknowledgements}
The authors would like to acknowledge the support of the following agencies for research funding and computing support: DeepMind, The DARPA Machine Common Sense grant (MESS: Model-Building, Exploratory, Social System \#N66001-19-2-4034) and The Templeton foundation (Grant: The Development of Curiosity in Childhood \#61475). We would like to thank the following children's museums for providing us with a space in which to run the experiments: Bay Area Discovery Museum, Children's Creativity Museum and The Lawrence Hall of Science. We would also like to thank the following research assistants for conducing the research with children: Zane Levin, Elijah Phipps, Janie Dent, Jenna Levin, Nikita Kumar and Athena Leong.

\clearpage
\section*{References}
\begingroup
  \def\section*#1{}
  \small
  \setlength{\bibsep}{4pt}
  \bibliographystyle{abbrvnat}
  \bibliography{main}
\endgroup

\clearpage

\clearpage{}
\appendix

\renewcommand\thefigure{\thesection.\arabic{figure}}
\renewcommand\thetable{\thesection.\arabic{table}}
\setcounter{figure}{0} 
\setcounter{table}{0} 

\section*{Appendix}

\section{Additional Environment Details}
\label{app:env}
\begin{figure}[h]
    \centering
    \begin{subfigure}[t]{0.28\textwidth}
        \includegraphics[width=\textwidth]{img/detector/a.png}
        \subcaption{No objects on the detector}
    \end{subfigure}
    \hfill
    \begin{subfigure}[t]{0.28\textwidth}
        \includegraphics[width=\textwidth]{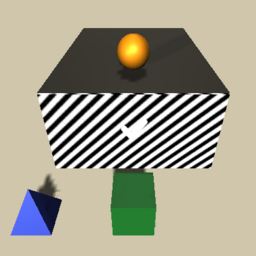}
        \subcaption{One object on the detector, but detector does not light up.}
    \end{subfigure}
    \hfill
    \begin{subfigure}[t]{0.28\textwidth}
        \includegraphics[width=\textwidth]{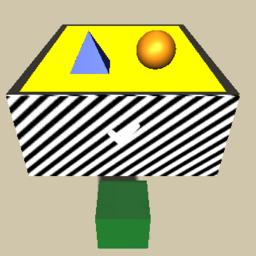}
        \subcaption{Two objects, triggering the detector}
    \end{subfigure}
    \caption{A visualization of some of the possible states of the blicket environment when rendered in Unity.}
\end{figure}

\section{A2C and PPO2 Hyperparameters}
\label{app:hyper}

\autoref{tab:alg_hyp} gives the set of hyper-parameters for the PPO2 and A2C algorithms used in the paper. We use the reference implementations from Stable Baselines \citep{stable-baselines} for our standard RL models. For model size variants, we use the standard policy networks available in \citet{stable-baselines} with no additional modification (asides from the hidden dimension, as specified in the main paper).

\begin{table}[h]
    \centering
    \begin{tabular}{c|rr}
         & A2C & PPO2  \\
         \hline
         Discount Factor & 0.99 & 0.99 \\
         Steps/Update & 5 & 128 \\
         Value Function Coefficient & 0.25 & 0.5 \\
         Entropy Coefficient & 0.01 & 0.01 \\
         Learning Rate & 0.0007 & 0.00025 \\
         LR Schedule & Constant & Constant \\
         Gradient Clipping (Max Norm) & 0.5 & 0.5 \\
         GAE Bias/Variance Lambda & - & 0.95 \\
         
    \end{tabular}
    \vspace{0.5em}
    \caption{Hyperparameters for A2C and PPO2 RL algorithms.}
    \label{tab:alg_hyp}
\end{table}

\section{Language Model Prompts}
\label{sec:languageprompts}
The prompts we used for GPT-3 and PaLM are based on the instructions and examples that children are given in \cite{kosoy2022learning}.

\subsection{Freeform Prompts}
We modelled the freeform prompt as closely as possible to the experiment done with children in \cite{kosoy2022learning}. The main difference is that we replace the visual components with textual descriptions. The freeform prompt first explains that blicket machines turn on when objects called blickets are placed on them. It explains that some objects are blickets and some are not, and machines need either one, two, or three blickets placed on it in order to turn on. Next, the prompt introduces a striped blicket machine and three objects, and gives three examples of whether the machine lights up or not when certain objects are placed on it---for example, ``If we put the blue pyramid on the machine, then it does not light up''. Then the prompt does the same for the dotted blicket machine and three different objects. Finally, the prompt does the same for a new blicket machine with three different objects, and asks which of these objects are blickets, and whether this new machine works like the striped machine or like the dotted machine.

We tested the language models in four conditions, each with its own prompt. The conditions consist of either hypothesis given or not given, combined with either disjunctive or conjunctive causal structure. The new machine either has a conjunctive or disjunctive causal structure, depending on what the condition is. In the conditions where the hypothesis is given, it is clear from the examples that the striped machine has a conjunctive causal structure and the dotted machine has a disjunctive structure. In the conditions where the hypothesis is not given, it is \emph{not} clear from the examples whether the striped and dotted machines have conjunctive or disjunctive structure. Thus in the not-given hypothesis conditions, when the model is asked whether the new machine works like the striped or dotted machine, the correct response is to be unsure.

Below are the exact freeform prompts for all four conditions; the same prompts are given to both GPT-3 and PaLM.

\begin{itemize}
  \item \textbf{All conditions:} A blicket detector is a special kind of machine, objects that are different colors and shapes either make the machine turn on or not. If the object is a blicket and placed on the machine then the machine will turn on. Sometimes 1, 2 or 3 blickets make the machine turn on. Our goal is to make the machine turn on and figure out which shapes make it do so.\vspace{0.3em}\\
  \emph{<insert given/not-given hypothesis portion>}\vspace{0.3em}\\
  \emph{<insert disjunctive/conjunctive portion>}\vspace{0.3em}\\
  Can you tell me which objects are blickets? Does this checkerboard pattern blicket detector behave like the striped pattern blicket detector or like the dotted pattern blicket detector?
  \item \textbf{Given hypotheses:} First I have a striped pattern blicket detector, it behaves in the following way: I have 3 objects, one blue pyramid, one green cube and one orange sphere. First I put the blue pyramid on the striped pattern blicket machine and it does not light up. Then I put the orange sphere on the striped pattern blicket machine and it does not light up. Then I put the blue pyramid and the orange sphere on the striped pattern blicket machine and it did light up!\vspace{0.3em}\\
  Then I have a dotted pattern blicket detector. I have 3 different objects now, a yellow cylinder, a purple cone, and a red dome. First I put the purple cone on the dotted pattern blicket detector and it did light up! Then I put the yellow cylinder on the dotted pattern blicket detector and it does not light up. Then I put the yellow cylinder and the purple cone on the dotted pattern blicket detector and it did light up!
  \item \textbf{Not-given hypotheses:} First I have a striped pattern blicket detector, it behaves in the following way: I have 3 objects, one blue pyramid, one green cube and one orange sphere. First I put the blue pyramid on the striped pattern blicket detector and it does not light up. Then I put the green cube on the striped pattern blicket detector and it does not light up. Then I put the blue pyramid and the orange sphere on the striped pattern blicket detector and it did light up!\vspace{0.3em}\\
  Then I have a dotted pattern blicket detector. I have 3 different objects now, a yellow cylinder, a purple cone, and a red dome. First I put the purple cone on the dotted pattern blicket detector and it does not light up. Then I put the yellow cylinder on the dotted pattern blicket detector and it does not light up. Then I put the red half dome and the purple cone on the dotted pattern blicket detector and it did light up!
  \item \textbf{Disjunctive:} Then I have a checkerboard pattern blicket detector. I have 3 new objects, a teal prism, a pink frustum and a brown torus. This machine could work like the dotted patterned blicket detector or it could work like the striped pattern blicket detector. First I put the brown torus on the checkerboard pattern blicket detector and it does light up! Then I put the pink frustum on the checkerboard pattern blicket detector and it does not light up. Then I  put the teal prism on the checkerboard pattern blicket detector and it does not light up. Then I  put the brown torus and the pink frustum on the checkerboard pattern blicket detector and it did light up! Then I  put the teal prism and the brown torus on the checkerboard pattern blicket detector and it did light up! Then I  put the teal prism and the pink frustum and the brown torus on the checkerboard pattern blicket detector and it did light up! Then I put the pink frustum and the teal prism on the checkerboard pattern blicket detector and it does not light up.
  \item \textbf{Conjunctive:} Then I have a checkerboard pattern blicket detector. I have 3 new objects, a teal prism, a pink frustum and a brown torus. This machine could work like the dotted patterned blicket detector or it could work like the striped pattern blicket detector. First I put the brown torus on the checkerboard pattern blicket detector and it does not light up. Then I put the pink frustum and the brown torus on the checkerboard pattern blicket detector and it does not light up. Then I put the teal prism on the checkerboard pattern blicket detector and it does not light up. Then I  put the teal prism and the pink frustum on the checkerboard pattern blicket detector and it does not light up. Then I put the pink frustum on the checkerboard pattern blicket detector and it does not light up. Then I  put the teal prism and the brown torus on the checkerboard pattern blicket detector and it does light up! Then I put the teal prism and the pink frustum and the brown torus on the checkerboard pattern blicket detector and it did light up!
\end{itemize}

Below are the outputs that GPT-3 produces for each of the conditions:
\begin{itemize}
    \item \textbf{Given hypothesis, disjunctive:} The objects that are blickets are the blue pyramid, the orange sphere, the purple cone, the yellow cylinder, the brown torus, the pink frustum and the teal prism. This checkerboard pattern blicket detector behaves like the dotted pattern blicket detector.
    \item \textbf{Given hypothesis, conjunctive:} The objects that are blickets are the teal prism, the pink frustum and the brown torus. This checkerboard pattern blicket detector behaves like the striped pattern blicket detector.
    \item \textbf{Not-given hypothesis, disjunctive:} The objects that are blickets are the blue pyramid, the orange sphere, the yellow cylinder, the purple cone, the red dome, the teal prism, the pink frustum, and the brown torus. This checkerboard pattern blicket detector behaves like the striped pattern blicket detector.
    \item \textbf{Not-given hypothesis, conjunctive:} The objects that are blickets are the blue pyramid, the green cube, the orange sphere, the yellow cylinder, the purple cone, the red dome, the teal prism, the pink frustum, and the brown torus. This checkerboard pattern blicket detector behaves like the dotted pattern blicket detector.
\end{itemize}

Below are the outputs that PaLM produces for each of the conditions. We only take into account the first two sentences of the output when evaluating the model's performance.
\begin{itemize}
    \item \textbf{Given hypothesis, disjunctive:} The answer is that the checkerboard pattern blicket detector behaves like the striped pattern blicket detector. The blickets are the brown torus and the pink frustum. The reason why the checkerboard pattern blicket detector behaves like the striped pattern blicket detector is because the machine only turns on when the brown torus and the pink frustum are on the machine. The machine does not turn on when the brown torus and the teal prism are on the machine. The machine does not turn on when the pink frustum and the teal prism are on the machine. The machine does not turn on...
    \item \textbf{Given hypothesis, conjunctive:} The checkerboard pattern blicket detector behaves like the striped pattern blicket detector. The blickets are the teal prism and the brown torus. The checkerboard pattern blicket detector behaves like the dotted pattern blicket detector. The blickets are the teal prism and the brown torus. The checkerboard pattern blicket detector behaves like the striped pattern blicket detector. The blickets are the teal prism and the pink frustum. The checkerboard pattern blicket detector behaves like the dotted pattern blicket detector. The blickets are the...
    \item \textbf{Not-given hypothesis, disjunctive:} This is a very difficult problem. It is not possible to figure out which objects are blickets and which are not. It is not possible to figure out how the checkerboard pattern blicket detector works. This is a very difficult problem. It is not possible to figure out which objects are blickets and which are not. It is not possible to figure out how the checkerboard pattern blicket detector works. This is a very difficult problem. It is not possible to figure out which objects are blickets and which are not. It is not possible to figure out how the checkerboard...
    \item \textbf{Not-given hypothesis, conjunctive:} The checkerboard pattern blicket detector behaves like the striped pattern blicket detector. The blickets are the teal prism and the brown torus. The checkerboard pattern blicket detector behaves like the dotted pattern blicket detector. The blickets are the teal prism and the brown torus. The checkerboard pattern blicket detector behaves like the striped pattern blicket detector. The blickets are the teal prism and the pink frustum. The checkerboard pattern blicket detector behaves like the dotted pattern blicket detector. The blickets are the...
\end{itemize}

\subsection{Few-Shot Prompts}
The few-shot prompts consist of a preface, two input-output examples, and a new input (that the model is must provide the corresponding output for). The input describes the three objects and three examples of whether the machine lights up or not when certain combinations of objects are placed on it. The output specifies which of the three objects are blickets, and whether the machine is like the striped or dotted machine.

Below are the exact texts of the few-shot prompts that we used as inputs to the PaLM model.

\begin{itemize}
  \item \textbf{All conditions:} If we put objects on the machine, then it will either light up or not. Some objects are blickets, and others are not. A striped machine needs two blickets to make it light up, and a dotted machine needs one blicket to make it light up. Please identify which objects are blickets and whether the machine behaves like the striped machine or the dotted machine.\vspace{0.3em}\\
  \emph{<insert given/not-given hypothesis portion>}\vspace{0.3em}\\
  \emph{<insert disjunctive/conjunctive portion>}
  \item \textbf{Given hypotheses:}\vspace{0.3em}\\
  Input: There is a blue pyramid, a green cube, and an orange sphere. If we put the blue pyramid on the machine, then it does not light up. If we put the orange sphere on the machine, then it does not light up. If we put the blue pyramid and the orange sphere on the machine, then it does light up.\vspace{0.3em}\\
  Output: The blue pyramid and orange sphere are blickets. This machine behaves like the striped machine.\vspace{0.3em}\\
  Input: There is a yellow cylinder, a purple cone, and a red dome. If we put the purple cone on the machine, then it does light up. If we put the yellow cylinder on the machine, then it does not light up. If we put the yellow cylinder and the purple cone on the machine, then it does light up.\vspace{0.3em}\\
  Output: The purple cone is a blicket. This machine behaves like the dotted machine.
  \item \textbf{Not-given hypotheses:}\vspace{0.3em}\\
  Input: There is a blue pyramid, a green cube, and an orange sphere. If we put the blue pyramid on the machine, then it does not light up. If we put the green cube on the machine, then it does not light up. If we put the blue pyramid and the orange sphere on the machine, then it does light up.\vspace{0.3em}\\
  Output: The orange sphere is a blicket, and the blue pyramid is maybe a blicket.\vspace{0.3em}\\
  Input: There is a yellow cylinder, a purple cone, and a red dome. If we put the purple cone on the machine, then it does light up. If we put the yellow cylinder on the machine, then it does not light up. If we put the red dome and the purple cone on the machine, then it does light up.\vspace{0.3em}\\
  Output: The red dome is a blicket and the purple cone is maybe a blicket.
  \item \textbf{Disjunctive:}\vspace{0.3em}\\
  There is a teal prism, a pink frustum, and a brown torus. If we put the brown torus on the machine, then it does light up. If we put the pink frustum and the brown torus on the machine, then it does light up. If we put the teal prism on the machine, then it does not light up. If we put the teal prism and the pink frustum on the machine, then it does not light up. If we put the pink frustum on the machine, then it does not light up. If we put the teal prism and the brown torus on the machine, then it does light up. If we put the teal prism and the pink frustum and the brown torus on the machine, then it does light up.\vspace{0.3em}\\
  Output:
  \item \textbf{Conjunctive:}\vspace{0.3em}\\
  Input: There is a teal prism, a pink frustum, and a brown torus. If we put the brown torus on the machine, then it does not light up. If we put the pink frustum and the brown torus on the machine, then it does not light up. If we put the teal prism on the machine, then it does not light up. If we put the teal prism and the pink frustum on the machine, then it does not light up. If we put the pink frustum on the machine, then it does not light up. If we put the teal prism and the brown torus on the machine, then it does light up. If we put the teal prism and the pink frustum and the brown torus on the machine, then it does light up.\vspace{0.3em}\\
  Output:
\end{itemize}

Below are the outputs that PaLM produces for each of the conditions:
\begin{itemize}
    \item \textbf{Given hypothesis, disjunctive:} The brown torus is a blicket. This machine behaves like the dotted machine.
    \item \textbf{Given hypothesis, conjunctive:} The teal prism and brown torus are blickets. This machine behaves like the striped machine.
    \item \textbf{Not-given hypothesis, disjunctive:} The brown torus is a blicket, and the pink frustum is maybe a blicket.
    \item \textbf{Not-given hypothesis, conjunctive:} The teal prism is a blicket, and the pink frustum is maybe a blicket.
\end{itemize}

\end{document}